\documentclass{article}
\usepackage{spconf,amsmath,graphicx}
\usepackage{epsfig}
\usepackage{amssymb}
\usepackage{booktabs}
\usepackage{color}

\usepackage[pagebackref=false,breaklinks=true,letterpaper=true,colorlinks,urlcolor=blue,citecolor=blue,linkcolor=blue,bookmarks=false]{hyperref}

\title{dual-modality seq2seq network for Audio-Visual event localization}
\name{Yan-Bo Lin\hspace{1pt}, Yu-Jhe Li\hspace{1pt}, and Yu-Chiang Frank Wang}
\address{
\hspace{1pt}Graduate Institute of Communication Engineering, National Taiwan University, Taiwan\\
\hspace{1pt}MOST Joint Research Center for AI Technology and All Vista Healthcare
}
\begin{document}
%
\maketitle
\begin{abstract}
Audio-visual event localization requires one to identify the event which is both visible and audible in a video (either at a frame or video level). To address this task, we propose a deep neural network named Audio-Visual sequence-to-sequence dual network (AVSDN). By jointly taking both audio and visual features at each time segment as inputs, our proposed model learns global and local event information in a sequence to sequence manner, which can be realized in either fully supervised or weakly supervised settings. Empirical results confirm that our proposed method performs favorably against recent deep learning approaches in both settings.
\end{abstract}
\section{Introduction}
\label{sec:intro}

While a substantial amount of visual and auditory signals can be observed in real-world activities, several studies in neurobiology and human perception suggest that the perceptual benefits of integrating visual and auditory information are extensive. For computational models, they reflect in lip reading~\cite{chung2017lip}, where correlations between speech and lip movements provide a strong cue for linguistic understanding. In sound synthesis~\cite{owens2016visually}, where physical interactions with different types of material give rise to plausible sound patterns. In spite of these advanced models, they are still limited to some constrained domains. Hence, some works~\cite{senocak2018learning,arandjelovic2017objects,Zhao_2018_ECCV,arandjelovic2017look} have begun to learn the correspondence between the visual scene and the sound, achieving the cross-modality scenario. However, such cross-modality learning methods typically assume that the audio and visual contents in a video are always presented and matched. Such assumptions might not be practical for analyzing/understanding unconstrained videos in the real world.

\begin{figure}[t]
  \begin{center}
  \includegraphics[width=0.95\linewidth]{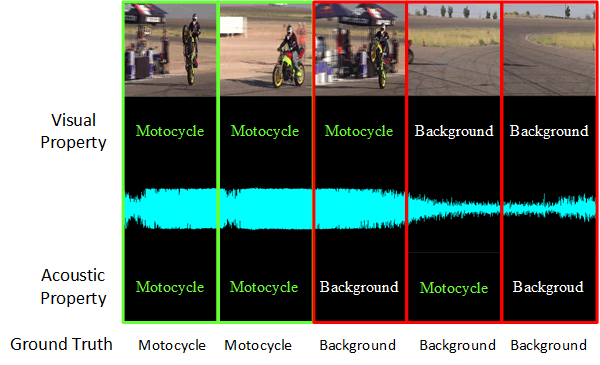}
  \vspace{-3mm}
  \caption{Illustration of audio-visual event localization. Note that green and red bounding boxes for video or audio data indicate the labels observed for single-modality data, while the ground truth labels are listed in the bottom row (i.e., non-background event labels are assigned only if cross-modality data exhibit same label information).}
  \label{fig:teaser}
  \end{center}
\end{figure}

Audio-visual event localization addresses the task of matching both visible and audible components in a video for identifying the event of interest, as depicted in Fig.~\ref{fig:teaser}. Recently, some multi-modal deep networks for joint audio-visual representation have been studied~\cite{hu2016temporal,yang2017deep,kiela2018efficient}. Aside from
works in representation learning, audio-visual cross-modal synthesis is studied in~\cite{zhou2017visual,chen2017deep}, and associations between natural image scenes and accompanying free-form spoken audio captions are explored in~\cite{harwath2016unsupervised}. However, how to handle learned audio and visual features for audio-video event jointly localization remains a challenging issue (i.e., hearing a dog barking and seeing a dog in the video simultaneously).
To this end, ~\cite{tian2018audio} introduce such novel problem of audio-visual event localization in unconstrained videos. As shown in Fig.~\ref{fig:teaser}, they define an audio-visual event as an event that is both visible and audible in a video segment. In addition, they further collect an Audio-Visual Event (AVE) dataset to systemically investigate three temporal localization tasks: supervised and weakly-supervised audio-visual event localization, and cross-modality localization. Nevertheless, they only consider temporal relationship within video or audio data, and no cross-modality relationship is observed.

\begin{figure*}[t]
  \centering
  \includegraphics[width=0.95\linewidth]{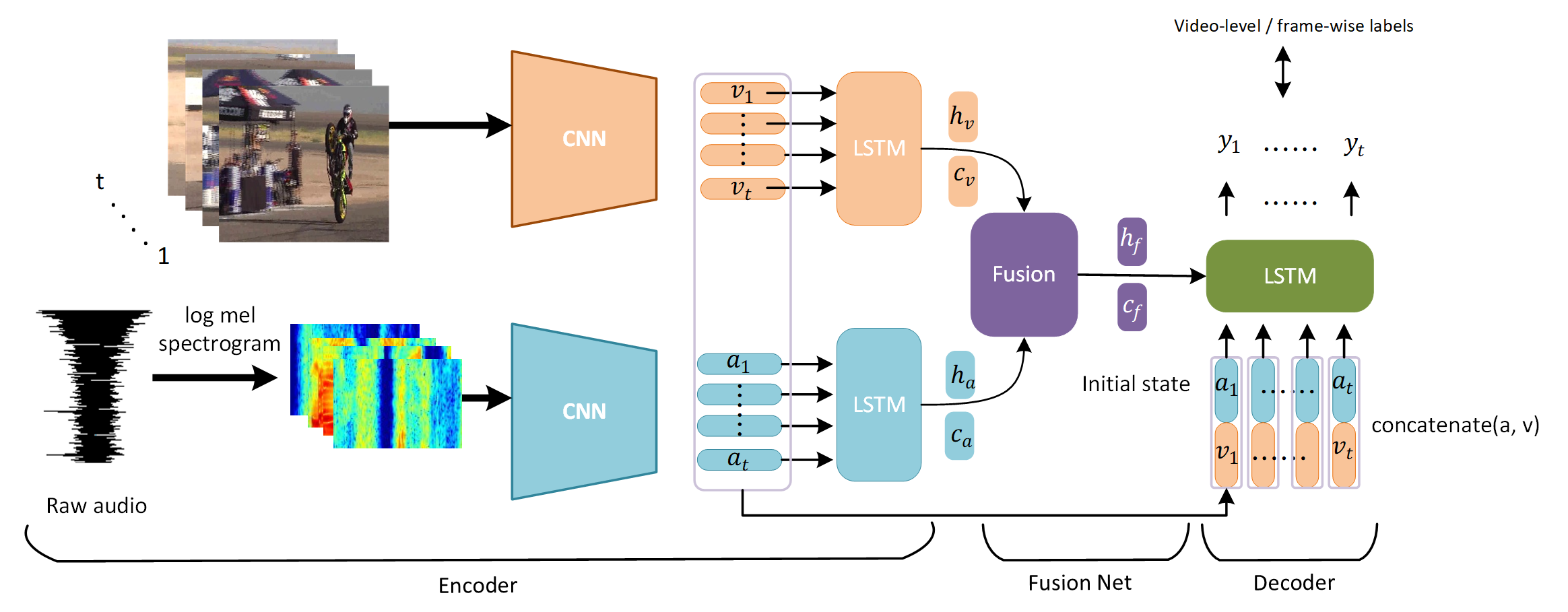}
  \caption{\textbf{Overview of our Audio-Visual Sequence-to-sequence Dual Network (AVSDN).} 
  Our AVSDN is composed of three main components, which include encoder modules (in orange and blue) for learning visual and audio representations, a fusion network (in purple) for producing global video representation, and a decoder (in green) which jointly takes global and local features for event localization. Note that $v$ and $a$ denote visual and audio features, $h$ and $s$ are the hidden and cell states of LSTMs, and $y$ indicates the event label.
  %
  }
  \label{fig:Model}
\end{figure*}

To address this challenging task, we propose an end-to-end deep learning framework of Audio-Visual sequence-to-sequence dual network (AVSDN). Based on sequence to sequence (seq2seq)~\cite{seq2seq} and autoencoder, our network architecture takes both audio and visual data at each time segment as inputs and exploits global and local event information in a seq2seq manner. More importantly, our model can be learned in a fully or weakly supervised settings, i.e., ground truth event labels observed in the frame or video levels. While the technical details of our proposed networks will be presented in Sec.~\ref{sec:method}, contributions of our work are highlighted as follows:
\begin{itemize}
  \vspace{-2mm}
  \item We propose a unique end-to-end trainable network for audio-visual event localization, which can be learned in fully or weakly supervised settings.
  \vspace{-2mm}
  \item Our proposed model jointly takes visual and audio data as inputs. By exploiting this cross-modality information across time, such encoded global features will be conditioned on the decoder for event localization.
  \vspace{-2mm}
  \item Experimental results demonstrate that our proposed model performs favorably against state-of-the-art approaches on the challenging AVE dataset.
\end{itemize}

\section{Proposed Method}
\label{sec:method}

\subsection{Problem Formulation}\label{sec:problem}

\noindent \textbf{Supervised event localization.} In supervised settings, one needs to predict video events every second given audio and visual tracks. As mentioned in Section~\ref{sec:intro}, non-background labels are determined only if visual and audio events are jointly observed. To this end, the segment-wised (every second) labels are given as $y_{t} =\left \{ y_{t}^{k}| y_{t}^{k}\in\left \{0,1 \right \},\sum\nolimits_{k=1}^{C+1} y_{t}^{k}=1 \right \}$, where $C$ denotes total event categories. The number of total categories is $C+1$ which includes one background label ,and $t$ is the time segment of one video. In supervised setting, segment-wise labels are observed during the training phase. 

\noindent \textbf{Weakly-supervised event localization.} Compared with the aforementioned supervised task, we can only access the video-level event labels in a weakly-supervised manner. The video-level event labels are processed by averaging the segment event labels $Y= \frac{1}{T}\sum\nolimits_{t=1}^T y_{t}$, where ${T}$ is the length of a video. For the weakly-supervised task, we can lower the reliance on the well-annotated labels and evaluate the robustness of our proposed framework.

\subsection{Audio-Visual Sequence-to-sequence Dual Network (AVSDN)}\label{sec:frame}

To address the audio-visual event localization problem in both supervised and weakly-supervised problems, we propose a novel framework named Audio-Visual sequence-to-sequence dual network (AVSDN). Based on  seq2seq~\cite{seq2seq} and autoencoder, our network architecture takes both audio and visual data at each time segment as inputs and exploits global and local event information in a seq2seq manner.

As shown in Fig~\ref{fig:Model}, the framework is composed of three components: encoder modules for learning visual and audio representations, fusion network for producing global video representation, and decoder to take global and local features for event localization. 
More details about the three components in AVSDN can be obtained as follows:
\\\\
\noindent \textbf{Encoder: learning global visual and audio representations.} The encoder, which is the first part of our network AVSDN, is aimed to extract global visual and audio representations for fusion network. 

Before learning the global features, we have to obtain the segment visual and audio representations by utilizing CNNs.
To better learn the visual and audio embedding features, our CNNs are learned from the large-scale dataset (ImageNet~\cite{ImageNet} and AudioSet~\cite{Audioset}) which are highly shown useful for vision and audition tasks. To be specific, for visual frames we sample a frame and obtain the visual representation from pre-trained ResNet-152~\cite{resnet}, which has been trained on ImageNet.
On the other hand, for one raw audio, we convert one segment to log mel spectrogram and extract an audio representation each 1s from VGGish~\cite{VGGish} trained on AudioSet.

In order to further learn the global visual and audio representations, we now utilize the Long Short-Term Memory (LSTM)~\cite{lstm}, which is known to exploit long-range temporal dependencies, to generate encoded temporal representation sequence. Generally, the most common implementation of vanilla LSTM~\cite{nLSTM} includes various gate mechanisms such as input gate, forget gate, output gate, memory state, and hidden state etc. 
%
The utilized LSTM unit in our proposed model is illustrated in Eq. (\ref{eq:lstm}).

\begin{equation}
\begin{gathered}
f_t = \sigma_g(W_fx_t + U_fx_t + b_f) \\ 
i_t = \sigma_g(W_ix_t + U_ix_t + b_i) \\
o_t = \sigma_g(W_ox_t + U_ox_t + b_o) \\
c_t = f_t\circ c_{t-1} + i_t\circ\sigma _c(W_cx_t + U_ch_{t-1} + b_c) \\
h_t = o_t\circ\sigma_h(c_t)
\end{gathered}
\label{eq:lstm}
\end{equation}

Each time step in Eq. (\ref{eq:lstm}) denotes subscript $t$, and $x_t\in \mathbb{R}^d$ denotes the given input of time step $t$ . $f_t\in \mathbb{R}^h$, $i_t\in \mathbb{R}^h$ and $o_t\in \mathbb{R}^h$ are forget, input and output gate's activation vector respectively. $h_t\in \mathbb{R}^h$ and $c_t\in \mathbb{R}^h$ are hidden and cell state of the LSTM unit. When $t=0$, $c_0 = 0 $ and $h_0 = 0 $ would be the initial values. $W\in \mathbb{R}^h\times d$, $U\in \mathbb{R}^h\times h$ and $b\in \mathbb{R}^h$ are weight matrices and bias vector which can be learned during train phase. Where $d$ and $h$ refer the number of input features and number of hidden units.
Element-wise product is denoted by $\circ$. Activation function: $\sigma_g$ is sigmoid function and $\sigma_c$ is hyperbolic tangent function. 

Eventually, all the audio and visual segments are the inputs of the two designed LSTM (audio and video separately). Hence, the last time step $T$ of hidden and cell state can be generated as the global representations of audio and visual tracks.
\\\\
\noindent \textbf{Fusion network: learning video event representation.} 

After obtaining the audio and visual global representations, our goal is to convert these two representations into one video event representation. To perform such a fusion mechanism, our fusion network is designed to fuse cross-modality features which are built based on dual multimodal residual network (DMRN) fusion block~\cite{tian2018audio}.
As mentioned above, the last time step $T$ of hidden and cell state from encoder can be given as the representations of audio and visual tracks. Following~\cite{tian2018audio}, with time step $T$, audio and visual hidden state ($h^a_T$, $h^v_T$) and cell state ($c^a_T$, $c^v_T$) can be fused with Eq.(\ref{eq:fusion}). After fusion hidden and cell state, these fused state will be the initial state of the decoder LSTM (one LSTM of right-half Fig.\ref{fig:Model}).
\begin{equation}
\begin{gathered}
h_T^{{a}\prime} = \sigma_c(h_T^a + \frac{1}{2}(g_\theta(h_T^a) + g_\theta(h_T^v)))\\
h_T^{{v}\prime} = \sigma_c(h_T^v + \frac{1}{2}(g_\theta(h_T^a) + g_\theta(h_T^v)))\\
h_T^{{f}} = h_T^{{a}\prime}  + h_T^{{v}\prime}\\
\end{gathered}
\label{eq:fusion}
\end{equation}
where $\sigma_c$ is denoted as hyperbolic tangent function and $g_\theta(.)$ is multilayer perceptrons (MLP) with parameters $\theta$.
The cell states can be fused like hidden states. The fused states, $h_t^{{f}}$ and $c_t^{{f}}$, turn to be the initial states in our decoder. Because compared with vanilla LSTM, a representative initial state can benefit a LSTM for prediction~\cite{seq2seq}. Thus, we take the fused states for the initialization for the decoder. 
\\\\
\noindent \textbf{Decoder: localization of video events using global and local cross-modality representations.} Generally, our decoder is aimed to perform the supervised and weakly-supervised event localization.
Thus, given both the fused global representations from the fused network and local features of audio and video, our decoder will generate the corresponding labels segment-wisely. The architecture of our decoder is a single LSTM. Different from each encoder, the inputs of the decoder are concatenated features which are global and local cross-modality representations.
We concatenate $a_t$ and $v_t$ which are audio and visual segment features from pre-trained CNN. Our decoder is designed to not only learn spatial cross-modality representations but temporal ones. Especially weakly supervised setting, we can only access to the video-level labels in the training phase. All the individual predictions will be aggregated by average pooling in Eq.(\ref{eq:avg}),
\begin{equation}
\begin{split}
\hat{m} = avg(m_1,m_2,...,m_T)= \frac{1}{T}\sum_{t=1}^{T} m_t,
\end{split}
\label{eq:avg}
\end{equation}
where $m_1,...,m_T$ are the predictions from the last fully connected layer of our model. The average prediction $\hat{m}$ over softmax function can be the probability distribution of the event category. For both the weakly-supervised and supervised setting, the predicted probability distribution can be optimized by video-level labels through binary cross-entropy.

\vspace{-2mm}
\section{experiments}
\label{sec:exp}
\subsection{Dataset}
Following~\cite{tian2018audio}, we consider \textit{Audio-Visual Event}(AVE)~\cite{tian2018audio} dataset (a subset of Audioset~\cite{Audioset}) for experiments. This AVE dataset includes 4143 videos with 28 categories and videos are labeled with audio-visual events every second. AVE dataset covers wide range domain events (e.g., Church bell, Dog barking, Truck, Bus, Clock, Violin, etc.).

\vspace{-3mm}
\subsection{Comparison results}
For both fully supervised and weakly-supervised audio-visual event localization, we apply \textbf{frame-wise accuracy} as an evaluation metric in both supervised and weakly-supervised settings. We compute the percentage of correct matchings over all the testing frames as prediction accuracy to evaluate the performance of audio-visual event localization.

\begin{figure}[t]
  \begin{center}
  \includegraphics[width=0.48\textwidth]{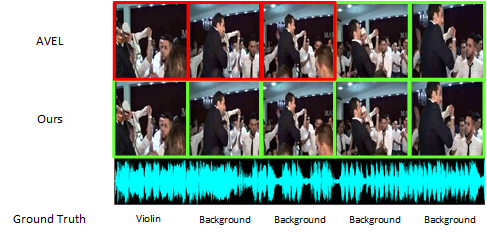}
  \vspace{-8mm}
  \caption{Visualization comparisons between AVEl and ours.}
  \label{fig:exp}
  \end{center}
\end{figure}

In this paper, we use different visual pre-trained embedding compared with Tian et al~\cite{tian2018audio}. Visual pre-trained embedding in Tian et al is VGG16~\cite{vgg}. Thus, we re-implement the model with ResNet-152~\cite{resnet} visual pre-trained embedding and show each one modality results. In a fully supervised manner, all the frame-wise labels are used during training. In Table \ref{table:STOA}, our model has better results compared with state-of-the-art methods even if the model~\cite{tian2018audio} is with cross-modality attention mechanism~\cite{att} which can find the audio location in the video scene~\cite{tian2018audio, arandjelovic2017objects}; In a weakly supervised manner, Table \ref{table:STOA_weak} shows our model outperforms other methods as well. Compared with the supervised task, our model can tickle noise labels better. We give an example of our visualization comparison with the existing method in Fig.~\ref{fig:exp}.

\begin{table}[t]
\caption{Comparisons with the state-of-the-art method of \cite{tian2018audio} in a \textbf{supervised} manner (all ground truth $y_{t}$ are observed during training. The number in bold indicates the best result.}
\vspace{2mm}
\resizebox{\linewidth}{!} {
\begin{tabular}{ccl}
\toprule
Method           & Accuracy (\%) & Remarks \\ 
\midrule
AVEL~\cite{tian2018audio}  &    59.5      &      audio only     \\ 
  &    55.3      &      visual only (VGG16)\\ 
  &    71.4      &     audio+visual (VGG16)       \\ 
 &    72.7      &     audio+visual w/ att (VGG16)  \\
  &    65.0      &      visual only (ResNet-152)      \\ 
  &    74.0      &     audio+visual (ResNet-152)       \\ 
  &    74.7      &     audio+visual w/ att (ResNet-152)     \\
\midrule
AVSDN      &     \textbf{75.4}     &   audio+visual (ResNet-152)        \\ 
\bottomrule
\label{table:STOA}
\end{tabular}
}
\end{table}
\vspace{-5mm}

\begin{table}[t]
\caption{Comparisons with the state-of-the-art method of \cite{tian2018audio} in a \textbf{weakly supervised} manner (only ground truth $Y$ is observed for training). The number in bold indicates the best result.}
\vspace{2mm}
\resizebox{\linewidth}{!} {
\begin{tabular}{ccl}
\toprule
Method           & Accuracy (\%) & Remarks \\ 
\midrule
AVEL\cite{tian2018audio}  &    53.4      &      audio only     \\ 
  &    52.9      &      visual only (VGG16)\\ 
  &    63.7      &     audio+visual (VGG16)       \\ 
  &    66.7      &     audio+visual w/ att (VGG16)  \\
  &    63.4      &      visual only (ResNet-152)      \\ 
  &    71.6      &     audio+visual (ResNet-152)       \\ 
  &    73.3      &     audio+visual w/ att (ResNet-152)     \\
\midrule
AVSDN      &     \textbf{74.2}    &   audio+visual (ResNet-152)       \\ 
\bottomrule
\label{table:STOA_weak}
\end{tabular}
}
\end{table}

\subsection{Ablation studies}

In this section, we will discuss the results on the different initial state for decoder LSTM. In Table \ref{table:ab}, first-two rows show the results which initial state is only from visual or audio content respectively. Further, we want to explore whether encoder LSTM can learn global event information or not.  With the last hidden states from visual and audio modality individually, these hidden states are guided by video-level labels through a simple multilayer perceptron (MLP). The third row in Table \ref{table:ab} shows the result is improved compared with only one modality. However, extra loss functions for guiding the last hidden states are not needed. The last hidden states are well-learned during the training of our AVSDN.

\vspace{-3mm}

\begin{table}[t]
\caption{Ablation studies on our network design, i.e., our decoder taking hidden and cell states of global visual/audio representations as conditioned LSTM inputs. Note that weakly supervised learning is considered in this table.}
\vspace{2mm}
\begin{center}
\resizebox{0.8\linewidth}{!} {

\begin{tabular}{lc}
\toprule
Method           & Accuracy (\%)  \\ 
\midrule
Visual input only   &   70.2          \\ 
Audio input only     &   70.9          \\ 
Audio + Visual (label guided)   &    72.6             \\ 
Fusion (Ours)     &    \textbf{ 74.2} \\
\bottomrule
\label{table:ab}
\end{tabular}
}
\end{center}
\end{table}

\section{conclusion}
\label{sec:cls}
\vspace{-2mm}

In this work, we proposed Audio-Visual sequence-to-sequence dual network (AVSDN) for video event localization, which can be learned in fully or weakly supervised fashions. Our network takes both audio and visual local features, together with integrated global representation, to perform event localization in a sequence to sequence manner. From the experimental results, the use of our network and its design can be successfully verified.

\noindent\textbf{Acknowledgements.}
This work is supported by the Ministry of Science and Technology of Taiwan under grant MOST 107-2634-F-002-010.



\bibliographystyle{IEEEbib}
\bibliography{strings,main}

\end{document}